\documentclass{article}
\pdfoutput=1
\usepackage{arxiv}

\usepackage[utf8]{inputenc} 
\usepackage[T1]{fontenc}    
\usepackage{hyperref}       
\usepackage{url}            
\usepackage{booktabs}       
\usepackage{amsfonts}       
\usepackage{microtype}      
\usepackage{xcolor}         
\usepackage{amsmath}
\usepackage{graphicx}
\title{Two out of Three (ToT): using self-consistency to make robust predictions}

\author{%
  Jung H. Lee \\
  Pacific Northwest National Laboratory\\
  Seattle, WA \\
  \texttt{jung.lee@pnnl.gov} \\
  \And
  Sujith Vijayan \\
  School of Neuroscience\\
  Virginia Tech\\
  Blacksburg, VA \\
  \texttt{neuron99@vt.edu} \\
}

\begin{document}

\maketitle

\begin{abstract}
Deep learning (DL) can automatically construct intelligent agents, deep neural networks (alternatively, DL models), that can outperform humans in certain tasks. However, the operating principles of DL remain poorly understood, making its decisions incomprehensible. As a result, it poses a great risk to deploy DL in high-stakes domains in which mistakes or errors may lead to critical consequences. Here, we aim to develop an algorithm that can help DL models make more robust decisions by allowing them to abstain from answering when they are uncertain. Our algorithm, named ‘Two out of Three (ToT)', is inspired by the sensitivity of the human brain to conflicting information. ToT creates two alternative predictions in addition to the original model prediction and uses the alternative predictions to decide whether it should provide an answer or not. 

\end{abstract}
\section{Introduction}
Deep learning (DL) can autonomously identify general rules from examples and use them to perform complex tasks \cite{Lecun2015, deep_review1, deep_review2}. DL models often outperform humans in a wide range of tasks. However, their automated solution discovery method does come with challenges. Unlike other computer programs, DL eliminates the need for engineered sets of instructions but it remains extremely difficult to determine DL’s decision making process or rectify errors because its operating principles are not explicitly specified by humans \cite{Lipton2016}.

As modern DL models contain numerous learnable parameters, far exceeding the number of examples on which they are trained (see \cite{Lecun1998},  for example), they are generally under-constrained. Additionally, given that DL is applied to domains with under defined and/or unpredictable problems, it is impractical to test DL models against every conceivable test case. Consequently, DL models will encounter edge cases and make unexpected mistakes that are difficult to detect or fix. This raises the question of whether current DL models should be deployed in high-stakes or safety-critical domains, in which errors may lead to catastrophic or fatal outcomes.

How do we ensure the safety of DL models in high-stakes problems? Humans may not be immune to making mistakes, but we can avoid making mistakes by learning to circumvent or stay away from decisions leading to critical consequences. Understanding the brain’s decision-making process has been challenging and is ongoing, but we note two observations. First, we evaluate the quality of our decisions; see \cite{conf1, conf2} for example. When we are confident about our decision, we are often correct. When we are not confident, we seek more information or decline to make a decision. Second, the brain is sensitive to conflicting information \cite{stroop}. For instance, it takes longer to read a word `blue’ written in red than to read ‘blue’ written in blue. The anterior cingulate cortex, a brain area thought to be important for high-level cognitive functions, is thought to function as a conflict detector \cite{acc}.  

Inspired by these observations, we hypothesize that the brain evaluates self-consistency to estimate the level of confidence in its decisions and determines whether or not to make a decision based on its confidence. If the relevant brain processes evaluating a circumstance reach an agreement, a decision can be made with confidence. If they disagree, more information or assistance is requested. With this hypothesis in mind, we develop a new algorithm called `Two out of Three (ToT)’, which enables DL models to make robust decisions using self-consistency. As the name implies, ToT creates two alternative predictions in addition to the original prediction and chooses its answer when two out of three predictions (i.e., original and at least one of the two alternative predictions) are consistent with each other. To create two alternative predictions, we adopt STCert proposed by an earlier study \cite{lee2023having} and derive a third prediction from hidden features. 

Here, we test whether ToT can help DL models make robust decisions using image classifiers. By robust decisions, we mean that DL models do not make predictions when in doubt and that they should produce highly accurate predictions if they are confident. As adversarial attacks pose great security challenges to DL \cite{reviewadver1, reviewadver2}, we test ToT using adversarial attacks in this study.

\section{Methods}
We describe below how ToT generates two alternative decisions.

\subsection{Second decisions from second views of inputs}
When we cannot identify a visual object in a scene, we take a closer look, thereby creating another physical view of the object. Moreover, the search-light hypothesis \cite{FRITZ2007} postulates that the brain can selectively choose relevant information from sensory inputs using selective attention, suggesting that this high-level cognitive function can create additional mental views of the object of interest. Thus, it may be the case that the brain uses multiple views of visual objects to make robust decisions. 

An earlier study \cite{lee2023having} proposed `Second Thought Certification (STCert)’, which mimics selective attention to confirm or reject DL models’ predictions. STCert uses an open-set segmentation model to detect regions of interest (ROIs) associated with the DL model’s original prediction; ROIs can be used as another view of the object associated with the original prediction. STCert compares an original prediction $P_{orig}$ (based on an image) with a second prediction $P_{2nd}$ (based on ROI) to confirm or reject the original prediction. Interestingly, when the original and second predictions are consistent, the original prediction is highly accurate. We adopt this STCert to create a second view of foreground objects in a scene. In the original version of STCert  \cite{lee2023having}, class labels were used as text prompts, but here we use superclass labels instead of class labels because the open-set segmentation model works better with coarse grain text prompts (e.g., dog) than with fine-grain text prompts (e.g., retriever).

In this study, we create 3 bounding boxes per ROI detected. The first box is identical to a ROI, and two bounding boxes contain additional background (i.e., spatial context); see Eqs. \ref{box1}-\ref{box3} and \cite{lee2023having}
 for more details. 
 \begin{equation}\label{box1}
     box1=ROI\equiv(x_1, y_1, x_2, y_2)
 \end{equation}
 \begin{equation}\label{box2}
     box2=(x_1-\delta, y_1-\delta, x_2+\delta, y_2+\delta)
 \end{equation}
 \begin{equation}\label{box3}
     box3=(x_1-2\delta, y_1-2\delta, x_2+2\delta, y_2+2\delta)
 \end{equation}
 where $\delta=5$. We resize all 3 bounding boxes to $(224,224)$ and obtain 3 predictions. That is, we have 3 second predictions for each ROI.
 
\subsection{Third decisions derived from hidden features}\label{third}
STCert compares original and second predictions to evaluate the confidence level of the original model, which would reflect the likelihood of their producing correct answers. However, when original and second predictions are different, STCert has no way of deciding upon the correct prediction. To address this issue, ToT derives a third prediction from hidden features of DL models. 

Inspired by earlier studies raising the possibility that features of the penultimate layer are well clustered according to labels of inputs \cite{deep_clus, deep_clus2, lee2024searchinginternalsymbolsunderlying}, we estimate functional clusters of hidden features in the penultimate layer and use them to map inputs onto functional codes. Then, we use these functional codes to infer the most probable predictions. Specifically, we adopt the method proposed by an earlier study \cite{lee2024searchinginternalsymbolsunderlying} and summarize it below:

\begin{enumerate}
  \item We extract hidden layer features $F(x,y)$ associated with ROIs, which encapsulate pixels of foreground objects. Spatial coordinates $x$ and $y$ depend on the resolution of feature maps. Regardless of the resolution of feature maps, we coarse-sample into $3 \times 3$ arrays. That is, $F(x,y) \rightarrow F_{x,y}$ , where $x,y=1,2,3$.
  \item We aggregate $F_{x,y}$ from all $n$ feature maps. That is, each input $x_m$ is converted into a 2-dimensional (D) array $\mathbf{FA}$ with shape ($9 \times n$). 
  \item We collect $\mathbf{FA}$ from training examples (200 per class). Then, we have $(9\times m, n)$ shaped array $\mathbf{A}$, where $m$ is the number of training examples.
  \item We remove the quiescent row (when all components are smaller than the mean values). 
  \item We standardize the 2D array,  $A_{ij}=(A_{ij}-\mu)/\sigma$ and use UMAP to extract the bases. 
  \item We use $k$-means clustering to create 1000 functional clusters $C_k$ of training examples, where $k=1,...,1000$. That is, each row $\vec{A_i}$ is mapped onto a cluster ($C_k$), $\vec{A_i} \rightarrow C_k$. It should be noted that each input can be mapped onto (maximally) 9 clusters, and 9 indices $(S_1, S_2, ..., S_9)$ of these clusters can effectively represent inputs. Consequently, these 9 indices are referred to as symbols. 
  \item Using the labels $l_m$ of training examples $x_m$, we create correlation map $CM(i, j)$ between $C_k$ and $l_m$; see Eq.\ref{eq_cmap}. 
  \begin{equation}\label{eq_cmap}
CM(i,j)=
 \begin{cases}
      CM(i, j)+1 & \text{if $i=S_k$, $j=l_m$}\\
      CM(i, j) & \text{otherwise}
    \end{cases}
\end{equation}
  \item  With functional clusters ($C_k$) , we calculate symbols $S_i$ ($0\leq S_k<1000)$ of any inputs $x$. In this study, we use $C_k$ and $CM(i,j)$ to convert testing examples (normal or adversarial) into 9 symbols $S_l$. We use 200 training examples randomly chosen from each class to calculate functional clusters ($C_k$) and correlation maps $CM(i,j)$. However, unlike the original study \cite{lee2024searchinginternalsymbolsunderlying} we use the entire pixel to obtain symbols of testing examples because we cannot use the ground truth labels of test examples to determine ROIs. We also find that ROIs of adversarial inputs are highly inaccurate. In brief, for training examples, $ROI \rightarrow  \vec{S} \equiv (S_1,...,S_9)$. For test inputs, $x_m \rightarrow  \vec{S}$. 
  \item  We  use $\vec{S}$ to calculate the probability of class $Pr_m$ for $x_m$ by using $CM(i,j)$. First, we calculate $P^o$ by taking SoftMax (Eq. \ref{eq_probability}).
\begin{equation}\label{eq_probability}
P(i=S_q,j)^o=\frac{e^{CM(i=S_q,j)}}{\sum_{n=1}^{cn} e^{CM(i=S_q,n)}} 
\end{equation}
, where $cn$ denotes the number of classes in the dataset ($cn=78$ or $64$ in our case). Second, as we obtain 9 $P^o$ for each $x_m$, we take average probability over $P^o$ (Eq. \ref{eq_average}) and then determine the most probable class as the prediction $Pr_m$ (Eq. \ref{eq_pred}).

\begin{equation}\label{eq_average}
P(j)^m=\frac{1}{9}\Sigma_{q=1}^9 P(S_q,j)^o
\end{equation}

\begin{equation}\label{eq_pred}
Pr_m=argsort_j P(j)^m
\end{equation}
\item We take top-2 predictions from $Pr_m$ and use them to determine the final answers.
  
\end{enumerate}

\subsection{Two out of Three}\label{tot}
\begin{figure}
  \centering
  \includegraphics[width=4.5in]{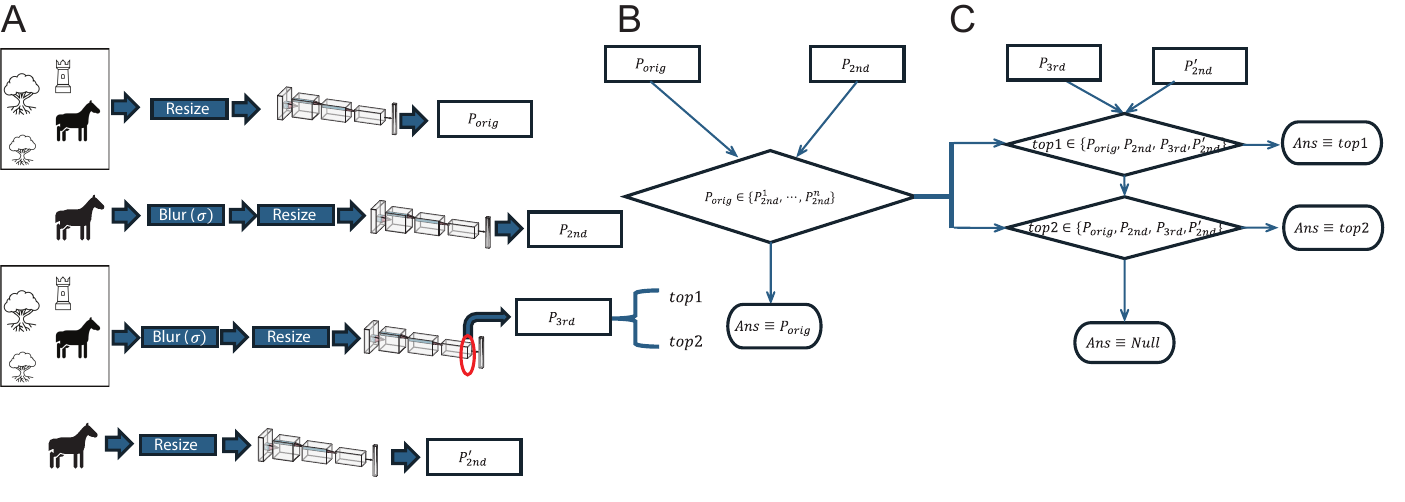}
  \caption{Overview of ToT. (A), 4 predictions used in ToT. In principle, ToT uses original, second and third predictions, but we find it is beneficial to use additional second predictions without blurring the images. Consequently, we have $P_{orig}$, $P_{2nd}$, $P_{3rd}$ and $P_{2nd'}$. (B), Comparing $P_{orig}$ and $P_{2nd}$. (C), Determine a final answer, which can be either `Null' or one of the top-2 predictions from symbols $\vec{S}$.}\label{diagram}
\end{figure}

ToT consists of two steps (Fig. \ref{diagram}). In the first step, it compares $P_{orig}$ and $P_{2nd}$ to determine if the original prediction is a high or a low confident answer. In the experiment, we evaluate ToT using adversarial examples and find that blurring inputs $x_m$ make the first step (i.e., comparing original and second predictions) effectively detect adversarial examples. Thus, ToT uses both blurred and original images in the first step. Specifically, ToT uses 4 different predictions (due to blurring), as shown in Fig. \ref{diagram}A. In the second step (\ref{diagram} B and C), ToT combines $P_{orig}, P_{2nd}, P'_{2nd}, P_{3rd}$ to make a final prediction (\ref{diagram} C). From the symbols ($\vec{S}$), we have top-2 predictions and compare each of them with $P_{orig},P_{2nd}, P'_{2nd}$.  If top-1 is consistent with either one, top-1 prediction is chosen as the final answer. If not, top-2 prediction is compared with original and second predictions. If top-2 are not consistent with original or second predictions, ToT does not provide the final answer (`Null' answer). Using this rule, ToT makes decisions only when at least two out of three predictions (original, second and third) are consistent to minimize its mistakes. 

\subsection{Dataset}
We use the two subsets (Mixed\_13 and Geirhos\_16) of ImageNet-1K \cite{imagenet} curated by a publicly available python robust machine learning library \cite{robustness, geirhos2020generalisationhumansdeepneural,santurkar2020breedsbenchmarkssubpopulationshift}.  Mixed\_13 contains 13 superclasses, each of which contains 6 fine-coarse classes; that is, there are 78 fine-grained classes. Geirhos\_16 contains 16 superclasses, each of which contains 4 fine-coarse classes. We choose the subset of ImageNet-1K for three reasons. First, most computer vision datasets contain a small number of classes, which can be grouped into super classes. As such, the two subset of ImageNets can approximate general computer datasets. Second, using the subset of ImageNet, we do not have to train new models. We can simply reuse a popular ImageNet model that has been well tested by the community. This eliminates the possibility of poorly trained DL models hindering fair evaluations of ToT. In this study, we remove all output nodes corresponding to classes, which are not included in the subset. Third, the subset of ImagNet, which has a moderate size, allows us to evaluate ToT efficiently and thoroughly with a limited computing resources. 

\subsection{Models}
To evaluate ToT’s capacity to help DL models make robust decisions, we use 4 convolutional networks (ResNet18, ResNet50, DenseNet121 and VGG19) \cite{resnet, vgg19, huang2018densely}  and 1 Vision Transformer (ViT) \cite{vit} pretrained on ImageNet. All models are implemented by using Pytorch \cite{Paszke2017} and Timms \cite{rw2019timm}, the open-source deep learning libraries. As stated above, we modify these models to carry subsets of output nodes, which does not influence their functionalities. For all our experiments, we use a consumer grade desktop equipped with core I9 CPU and RTX4090 GPU. Its CPU and GPU have 64 GB and 24 GB memory, respectively. 

\subsection{Adversarial attacks}

Two studies \cite{szegedy2014intriguingpropertiesneuralnetworks, goodfellow2015explainingharnessingadversarialexamples} showed that imperceptible perturbations can easily manipulate DL models’ decisions, and since then, numerous studies proposed adversarial attacks on DL models, confirming that DL models are vulnerable to various types of adversarial attacks; see \cite{reviewadver1, reviewadver2} for a review. Despite their diverse characteristics, they fall into two categories, white box and black box attacks, depending on adversaries’ access to models. White box attacks require full access to models and their gradient signals. Using the gradients, modifying inputs to disrupt DL models’ decisions can be straightforward, but even a tiny perturbation can disrupt DL models’ operations. Black box attacks rely on queries and DL models’ answers to approximate gradient signals and use approximations to craft adversarial attacks. The most common method to approximate the gradients is to use queries. 

Adversarial attacks remain one of the most significant security threats for DL models. When DL models are trained to work on safety critical problems, they would become high-value targets. Thus, it is imperative that we find ways to protect DL models from adversarial attacks before they are deployed. In this study, we ask if ToT can help DL models make robust decisions in the presence of adversarial attacks. Specifically, we use `Projected Gradient Descent (PGD)' \cite{pgd} and `AutoAttack' \cite{autoattack} to craft adversarial inputs, using an ensemble of 4 attacks (the two variants of protected gradient descents, FAB and square attacks). The adversarial inputs are created for all 5 models respectively and used to evaluate the efficacy of ToT. For both attacks, we use $L_{\infty}=0.03$. After crafting the adversarial examples using a public python toolbox, AdverTorch \cite{ding2019advertorch} and codes provided by the authors of AutoAttack \cite{autoattack}, we save them as 8-bit images and use them to evaluate ToT.

\section{Results}
ToT uses both confidence estimates and self-consistency to allow DL models to make robust decisions. By “robust decisions”,  we mean that DL should avoid making mistakes while presenting highly accurate predictions. Consequently, ToT has two aims. First, DL models should be able to refuse to produce predictions when they are not confident. Second, their decisions should be highly accurate when they are confident, which would lead them to make predictions. Here, we apply ToT to image classifiers and examine if it can make image classifiers robust. ToT first evaluates confidence level by utilizing the second views of foreground objects (Fig. \ref{diagram}A and B) and uses three decisions to produce final answers (Fig. \ref{diagram}C). We ask if ToT can make robust predictions on adversarial inputs, the most serious security threat to DL models. 

Below, we discuss our empirical evaluation of ToT by using image classifiers. 

\subsection{Confirm or reject the original prediction by using the second prediction}

ToT creates a second view of the same image. Specifically, we use an open-set segmentation model \cite{liu2023grounding} to detect ROIs and use it as a second point of view. After asking the same model to predict the label of ROIs, there are two predictions, an original prediction on an entire image and a second prediction on ROIs. ToT compares them to determine the confidence level of the original prediction. 

Although we test DL models, pretrained on ImageNets, we remove output nodes corresponding to the classes that are not included in the subsets. The removal of output nodes does not fundamentally change their properties, but their answers will be different due to the output nodes becoming smaller. For example, the chance level increases from $1/1000$ to $1/78$ in Mixed\_13. As such, we first evaluate the accuracy of 5 DL models on the two subsets, Mixed\_13 and Geirhos\_16 (see supplementary Table \ref{table}). Then, we test how well the second prediction can help DL models make robust decisions. Specifically, we ask how often DL models would reject original predictions and evaluate the accuracy of the certified predictions, to which original and second predictions would agree. In the first step, it obtains second predictions by identifying ROIs via the semantic segmentation model, using the super class name of the original prediction, which is readily available in the two subsets. In the original STCert, the fine-grained class name is used, but we find that the super class name produces better results. In the second step, we ask the same DL models for their second predictions on ROIs  (i.e., the regions encapsulating the foreground objects). 

\begin{figure}
  \centering
  \includegraphics[width=5.5in]{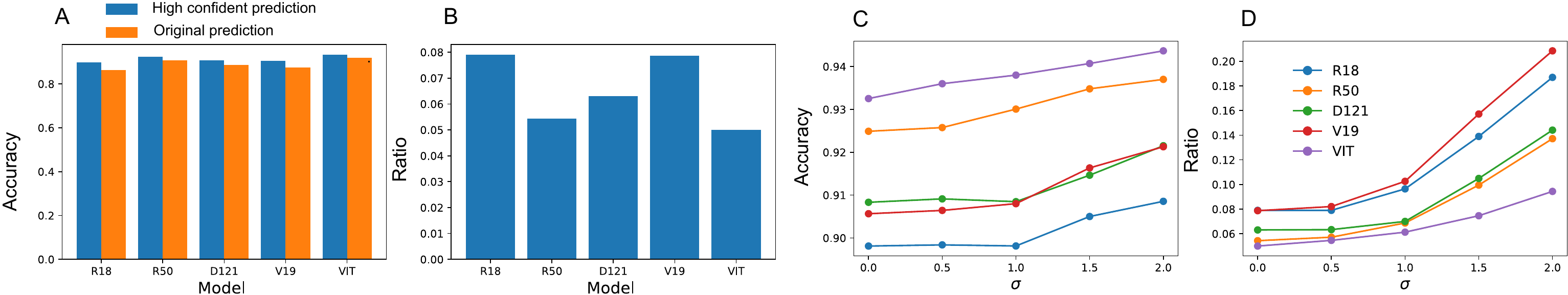}
  \caption{(A), Accuracy of high confident predictions on Mixed\_13. The blue bars denote the accuracy of high confident predictions, whereas the orange bars denote the accuracy of original predictions. (B), Rate of low confident predictions (i.e., the number of inputs, which original and second predictions disagree). (C), Accuracy of high confident predictions depending on the kernel size ($\sigma$) of Gaussian blur kernel. (D), Rate of low confident predictions depending on the kernel size ($\sigma$) of Gaussian blur kernel. R50, V19, D121 and VIT denote ResNet18, ResNet50, VGG19, DenseNet121 and Vision Transformer.}\label{Mixed}
\end{figure}

\begin{figure}
  \centering
  \includegraphics[width=5.5in]{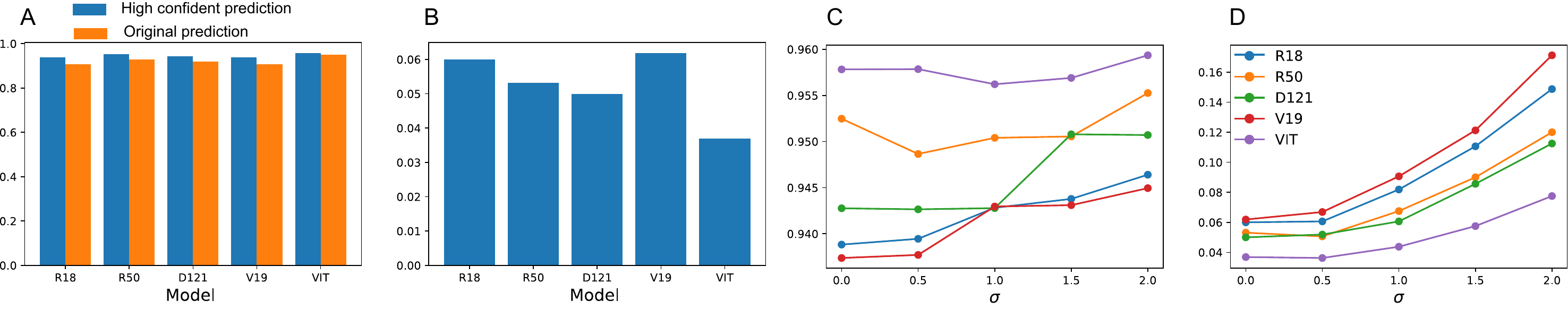}
  \caption{(A), Accuracy of high confident predictions on Geirhos\_16. The blue bars denote the accuracy of high confident predictions, whereas the orange bars denote the accuracy of original predictions. (B), Rate of low confident predictions (i.e., the number of inputs, which original and second predictions disagree). (C), Accuracy of high confident predictions depending on the kernel size ($\sigma$) of Gaussian blur kernel. (D), Rate of low confident predictions depending on the kernel size ($\sigma$) of Gaussian blur kernel.}\label{Geirhos}
\end{figure}

We note that there are 3 second predictions (see Eqs. \ref{box1}-\ref{box3} and related text). Additionally, a segmentation model often returns multiple ROIs because some images contain multiple objects in the same super classes or they identify multiple ROIs in error. Thus, we aggregate all second predictions available and test if the original prediction is included in the set of second predictions. If the original prediction is included in the set of second predictions, it returns the original prediction as the high-confident prediction. If not, we consider the prediction as one of the low-confident answers. Figs. \ref{Mixed} and \ref{Geirhos} summarize our evaluation of this approach using Mixed\_13 and Geirhos\_16, respectively. As shown in the figure, the high confident predictions are more accurate than the original predictions alone (Figs. \ref{Mixed}A and \ref{Geirhos}A), but the higher level of accuracy of the certified predictions does come with consequences. The model refuses to make predictions on $\approx 6\%$ inputs (Figs. \ref{Mixed}B and \ref{Geirhos}B). 

The ability (or the choice) to refuse to provide an answer may be worth having in high-stakes domains, where incorrect predictions cause more harm than nullified predictions (i.e., refusal to answer). This leads us to ask if we can further improve the accuracy of high confident predictions. Interestingly, if ROIs are blurred, the accuracy of high-confident predictions increases. Figs. \ref{Mixed}C and \ref{Geirhos}C show the accuracy of high confident predictions depending on the kernel of Gaussian blur filter \cite{clark2015pillow}. Notably, the number of inputs, on which DL models refuse to make predictions, also increases (Figs. \ref{Mixed}D and \ref{Geirhos}D). That is, the second predictions on blurred ROIs can make the accuracy of high confident prediction higher by rendering DL models to be more careful in making predictions. 

\subsection{Two out of Three (ToT): Two are needed to answer}
The results above suggest that second predictions especially with blurred ROIs can assist DL models to make robust decisions on normal inputs. Then, can we use this approach to prevent DL models from making incorrect predictions on adversarially manipulated inputs that pose great security risks to DL? To address this question, we craft adversarial perturbations for all 3900 validation images of Mixed\_13 and 1600 validation images of Geirhos\_16 and estimate how often the original and second predictions would differ depending on the kernel size of Gaussian blur filter. That is, we estimate the ratio of low-confident predictions on adversarial images. 

\begin{figure}
  \centering
  \includegraphics[width=4.0in]{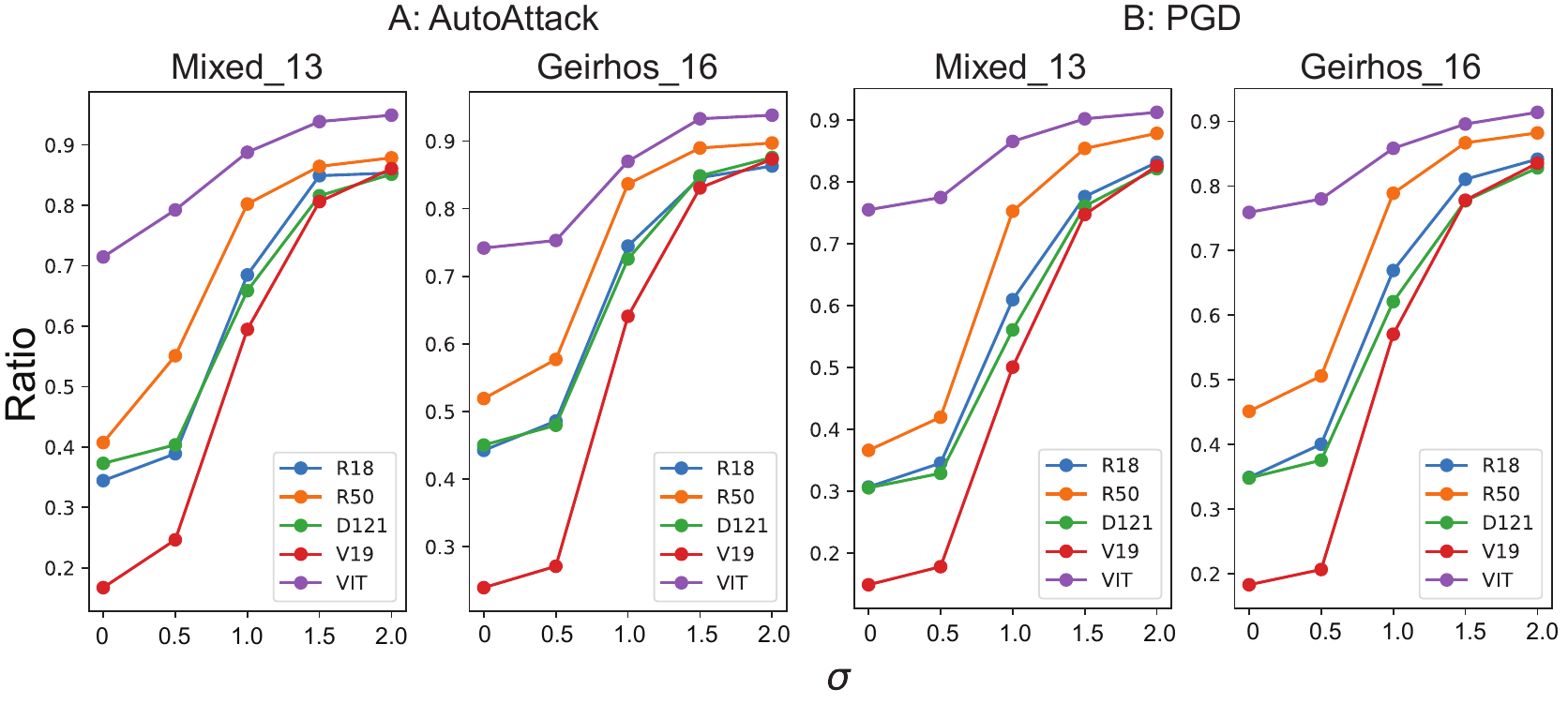}
  \caption{Ratio of low-confident predictions on adversarial inputs depending on the kernel size ($\sigma$) of Gaussian blur kernel. (A), Low-confident prediction ratio when auto-attack is used to craft adversarial inputs. (B), Low-confident prediction ratio when PGD is used. This ratio is equivalent to the detection rate of adversarial attacks in ToT. }\label{detect}
\end{figure}

As shown in Fig. \ref{detect}, the ratio of low-confident predictions strongly depends on the $\sigma$. As $\sigma$ increases, the ratio increases rapidly. Around at $\sigma=1.5$, the ratio becomes higher than 80\%. If we reject the low-confident predictions, DL models can avoid making mistakes on adversarial inputs. Now, the question is, which prediction is the correct one, when the original and second predictions are different? Alternatively, when original predictions are considered low confident, how do we reach the final prediction or answer?

Clearly, additional information is necessary to make a final decision, and thus, ToT derives a third decision from hidden features (section \ref{third}) and uses them to select a final decision (section \ref{tot}). ToT consists of two stages. In the first stage, ToT compares the original and second predictions. An original prediction is made on a present image, and a second prediction is made on a blurred ROI (Fig. \ref{diagram}B), since it can improve the accuracy of the high-confident predictions (Figs. \ref{Mixed}C and \ref{Geirhos}C). Without blurring ($\sigma=0$), the second prediction is often consistent with the original prediction modified by adversarial examples (see the ratio at $\sigma=0$ in Fig. \ref{detect}). In the second stage, ToT uses all three (original, second and third) predictions. Specifically, using the correlation maps (Eq. \ref{eq_cmap}) between hidden features projected to UMAP space and labels of inputs, we infer top-2 predictions (Fig. \ref{diagram}C) and test if one of them is consistent with the original prediction on an entire input or the second prediction on ROI. When top-1 prediction is consistent with one of the two predictions, top-1 prediction is the final answer. If top-1 prediction is inconsistent, top-2 prediction is compared. If all top-2 predictions are inconsistent with one of the predictions, our algorithm refuses to make any prediction. 

\begin{figure}
  \centering
  \includegraphics[width=4.5in]{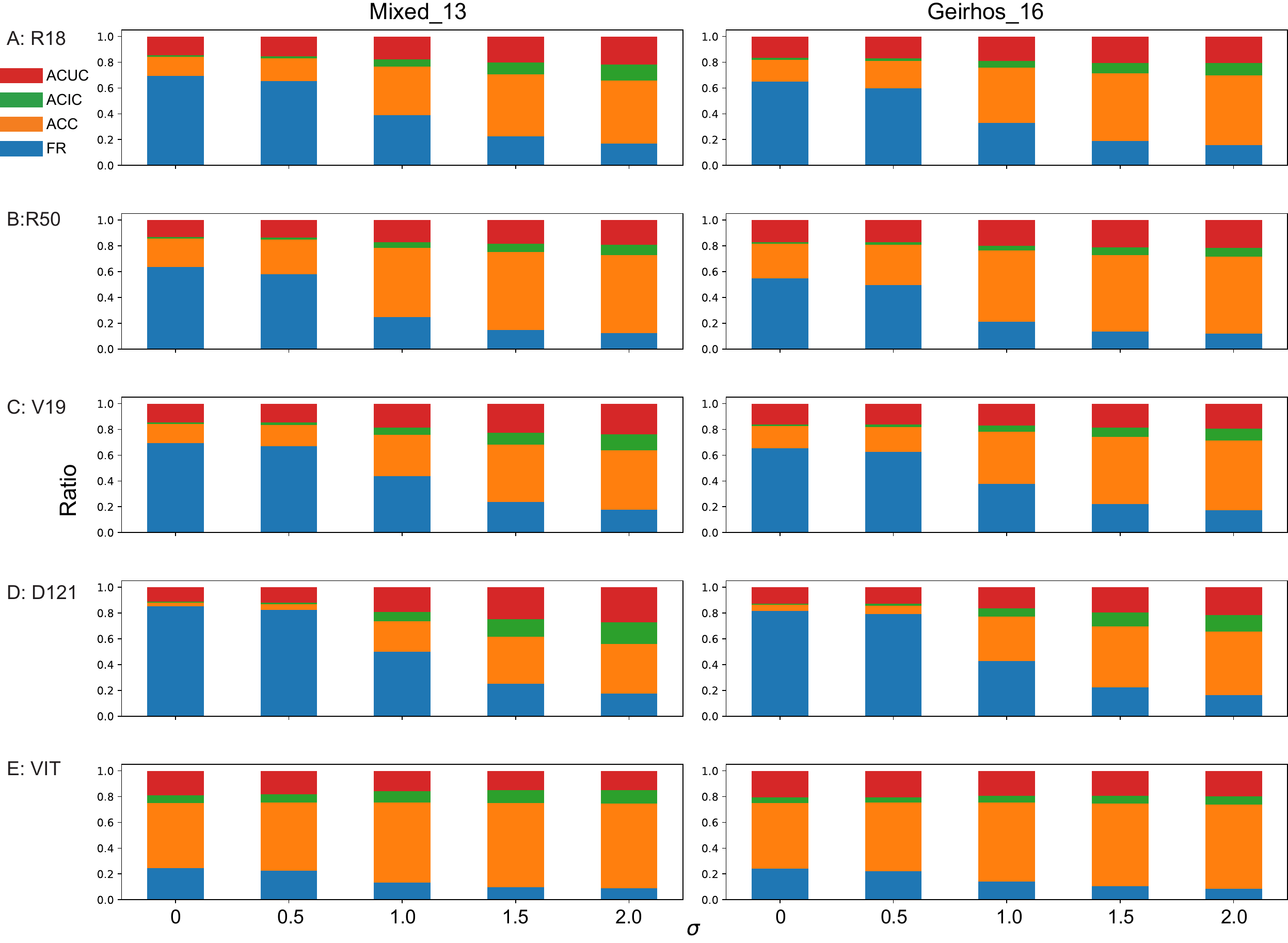}
  \caption{Evaluation of ToT on adversarial inputs crafted by PGD. (A)-(E), evaluation of ResNet18, ResNet50, VGG19, DenseNet121 and ViT, respectively. We estimate the failure ratio (FR) of detection, the accuracy of final answers  (ACC), the rate of inaccurate final answers (ACIC) and the rate of inputs (ACUC), on which ToT refuses to make predictions. They are shown in blue, orange, green and red, respectively. We show evaluation on Mixed\_13 in the left column and those on Geirhos\_16 in the right column. }\label{pgd}
\end{figure}

\begin{figure}
  \centering
  \includegraphics[width=3.5in]{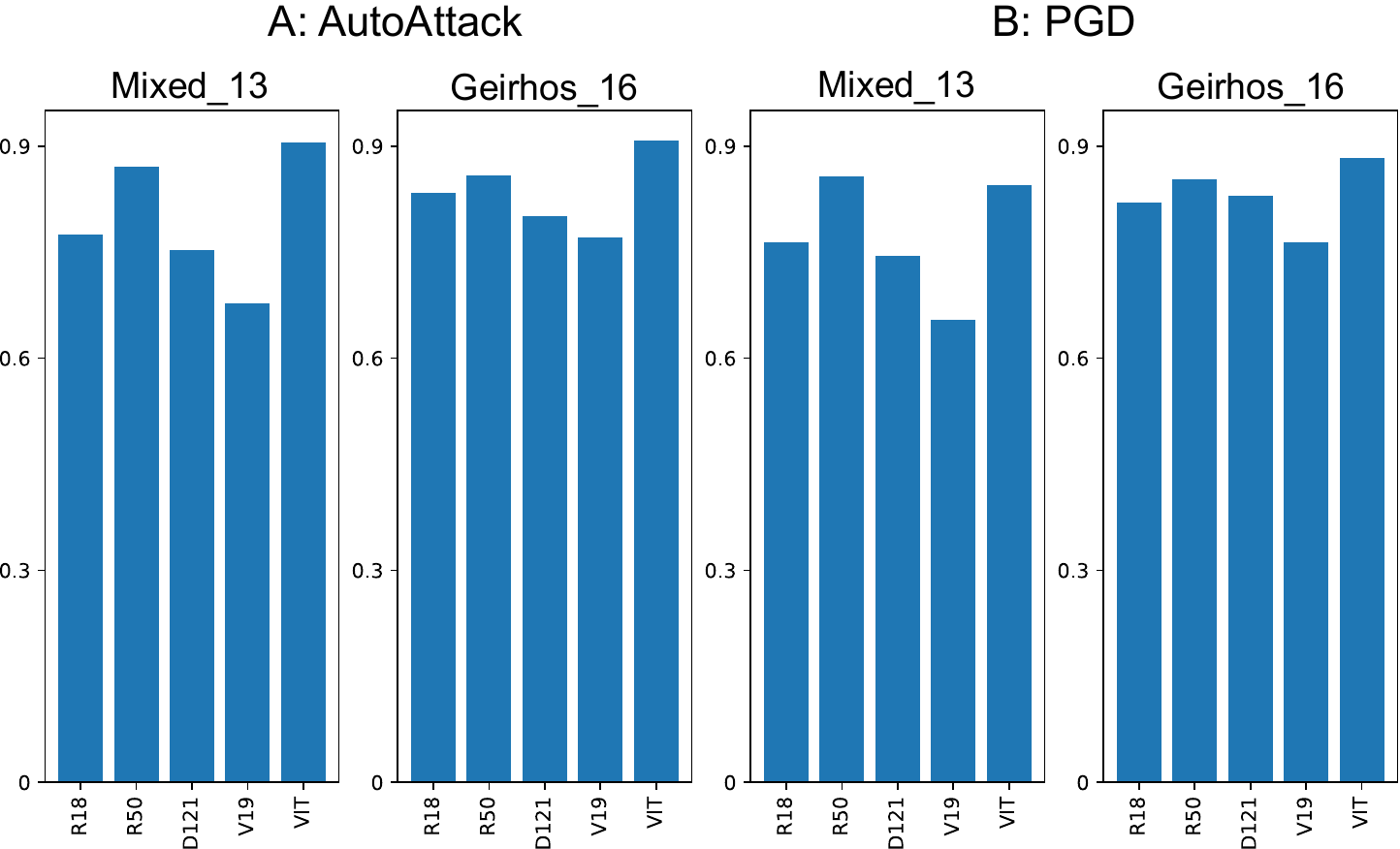}
  \caption{Accuracy of ToT's final answers on adversarial inputs. (A), Accuracy on inputs created by AutoAttack, (B), Accuracy on inputs created by PGD. We evaluate them for Mixed\_13 and Geirhos\_16, respectively. As ToT is allowed to refuse to make predictions, the accuracy is calculated using final predictions different from `Null'. }\label{final_model_comp}
\end{figure}

Fig. \ref{pgd} and Supplementary Fig. \ref{autoattack} summarize our evaluation. The blue bars denote how often ToT classifies adversarial inputs as high-confident predictions. As all inputs are adversarial inputs, they are equivalent to the ratio of detection failure. For the rest of the inputs, ToT tries to provide final answers, which are not necessarily different from low-confident original answers. ToT's final decisions on low-confident predictions can be correct (ACC, orange bars in the figure) or incorrect (ACIC, green bars). Alternatively, ToT can refuse to provide final answers (ACUC, red bars). Across 5 models, 2 datasets and 2 attack types, we observe that ACC rapidly increases as $\sigma$ increases, but ACIC and ACUC do not increase rapidly. This means that ToT can detect adversarial inputs and even correct original predictions manipulated by adversarial attacks. Thus, we further evaluate the accuracy of ToT's final answers, when $\sigma=2.0$. As shown in Fig. \ref{final_model_comp}, when ToT returns final answers, their accuracy is high (60-90\%). We observe that VGG19 shows the lowest level of accuracy, whereas VIT shows the highest level of accuracy. 

The results above suggest that ToT can strengthen DL models and lead them to make robust decisions even on adversarial inputs. Finally, we test the full version of ToT using clean inputs. We split predictions into 5 different categories, high confident correct (HCC, blue bars), high confident incorrect (HCIC, orange bars), low confident correct (LCC, green bars), low confident incorrect (LCIC, red bars) and Null (LCUC, purple bars); see Supplementary Fig. \ref{norm}. 

\begin{figure}
    \centering
    \includegraphics[width=5.0in]{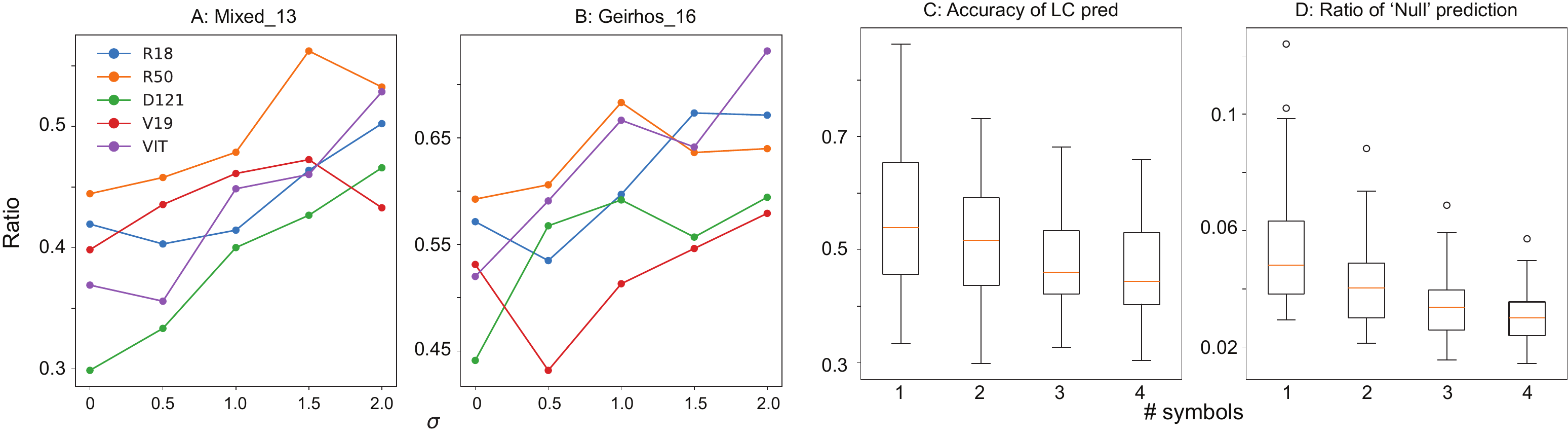}
    \caption{(A), Accuracy of LCC depending on Gaussian blur kernel size $\sigma$. (B), Ratio of LCUC, (C), Accuracy of LCC depending on the number of top predictions from symbols. In this plot, we aggregate the accuracy from all 5 models, 2 datasets (Mixed\_13, Geirhos\_16) and 2 attack types (AutoAttack and PGD). (D), Ratio of LCUC depending on the number of top predictions.}
    \label{comp-norm}
\end{figure}

ToT further assesses low confident answers and makes final predictions. We evaluate the accuracy of final predictions on low-confident inputs and find that the accuracy goes up, as $\sigma$ becomes bigger (Fig. \ref{comp-norm}A), but the ratio of Null prediction also goes up (Fig. \ref{comp-norm}B). Furthermore, the accuracy of final predictions on low confident inputs can be controlled by the number of predictions from symbols. The more predictions are used in ToT, the worse the accuracy of final predictions becomes (Fig. \ref{comp-norm} C), but the ratio of Null prediction declines (Fig. \ref{comp-norm} D).

\section{Discussion}
In this study, we propose ToT that allows DL models to use self-consistency to evaluate original predictions and correct them when necessary. Our evaluation suggests that ToT can help DL models avoid mistakes by allowing them to refuse to make predictions on uncertain inputs. 

Below, we discuss how ToT is related to earlier works and its limitation. 
\subsection{Links to earlier works}
DL models are sensitive to natural data shift and perturbations (both adversarial and non-adversarial), leading them to make unstable (non-robust) predictions. Many studies sought effective algorithms that could make DL models’ predictions more stable over data shift and perturbations. Drenkow et al. \cite{drenkow2022systematic} reviewed and classified them into three tactics. The first tactic searches for the architectures that can make DL more robust.  A few studies  \cite{arch_robust2, arch_robust2, LN-NLP} proposed preprocessing modules that denoise inputs. One of these modules adopts the linear-nonlinear-Poisson (LNP) model developed to model primary visual cortex \cite{LN-NLP}. The second tactic probes the effects of data augmentation on DL models’ robustness. Notably, if DL models are trained to produce invariant answers over a specific corruption/noise, they become robust to chosen corruptions \cite{geirhos2020generalisationhumansdeepneural}. Conversely, data augmentation can make DL models more robust. The third tactic directly trains DL models to be robust to perturbations of interest. The most notable example is adversarial training \cite{shafahi2019adversarialtrainingfree}. By using Min-Max algorithm, DL models can be trained to be robust to adversarial examples. Although adversarial training is considered as one of the strongest attacks, it is expensive and known to be attack-specific \cite{bai2021recent, LN-NLP}. All three tactics may improve the robustness of DL models under certain conditions, but we still do not know how to build fully reliable and robust architectures, how to choose proper data augmentation and what perturbations to use during adversarial training. 

The earlier studies mentioned above indicate that DL models’ sensitivities to data shift or perturbations are fundamental properties of DL, which would be challenging to address. Inspired by the fact that humans decline to answer difficult questions to avoid critical errors, we propose ToT, an algorithm that allows DL models to evaluate levels of confidence and refuse to provide answers when they are uncertain. As ToT creates two additional predictions and combine them with the original prediction, it is implicitly related to ensemble learning \cite{Hertz}. Unlike traditional ensemble learning, however, ToT derives additional predictions from the same model. We also note 1) ToT uses additional modules to extract internal representations, which is related to the first tactic mentioned above and 2) ToT blurs inputs, previously used for cleansing of adversarial attacks \cite{shah2023trainingfoveatedimagesimproves, gauss_blur, gauss_blur2}, which is related to the second tactic mentioned above. 
\subsection{Limitations}
The aim of this study is to test if self-consistency (regarding multiple points of view) can be used to prevent DL models from making mistakes. To this end, we design ToT and test it with image classifiers. While the same principle can be applied to a wide range of models and domains, testing self-consistency explicitly and confirming whether it could be used in more general cases may be necessary. We further note that multiple and different methods can be used to implement ToT. For instance, preprocessing, different from Gaussian blur, could have been used and outperform Gaussian blur used in ToT. Similarly, decisions from hidden layer features can be derived in more efficient ways. We plan to address these limitations in the future.

\subsection{Broader Impacts}
We propose a potential algorithm that makes DL models' decision more robust, which will help us reduce DL models’ mistakes when deploying DL systems in safety-critical domains. We are not aware of any potential negative impacts.
\bibliography{ref}

\bibliographystyle{unsrt}

\appendix

\setcounter{table}{0}
\renewcommand{\tablename}{Supplementary Table}

\setcounter{figure}{0}
\renewcommand{\figurename}{Supplementary Figure}
\section{Appendix: Supplementary Figures and Table}

\begin{table}[h]
\caption{Accuracy of DL models on the two subsets of ImageNet. It should be noted that the accuracy is measured after the output nodes, which correspond to classes not included in the two subsets, are removed.} \label{table}
\begin{center}
\begin{tabular}{|ccc|}
\hline

Model & Mixed 13 & Geirhos 16 \\ \hline
R18   & 86.3 \%    & 90.8 \%      \\
R50   & 90.6 \%    & 92.9 \%      \\
V19   & 87.4 \%    & 90.8 \%      \\
D121  & 88.5 \%    & 90.2 \%      \\
VIT   & 91.9 \%    & 95.1 \%      \\ \hline
\end{tabular}
\end{center}
\end{table}

\begin{figure}[h]
  \centering
  \includegraphics[width=5.0in]{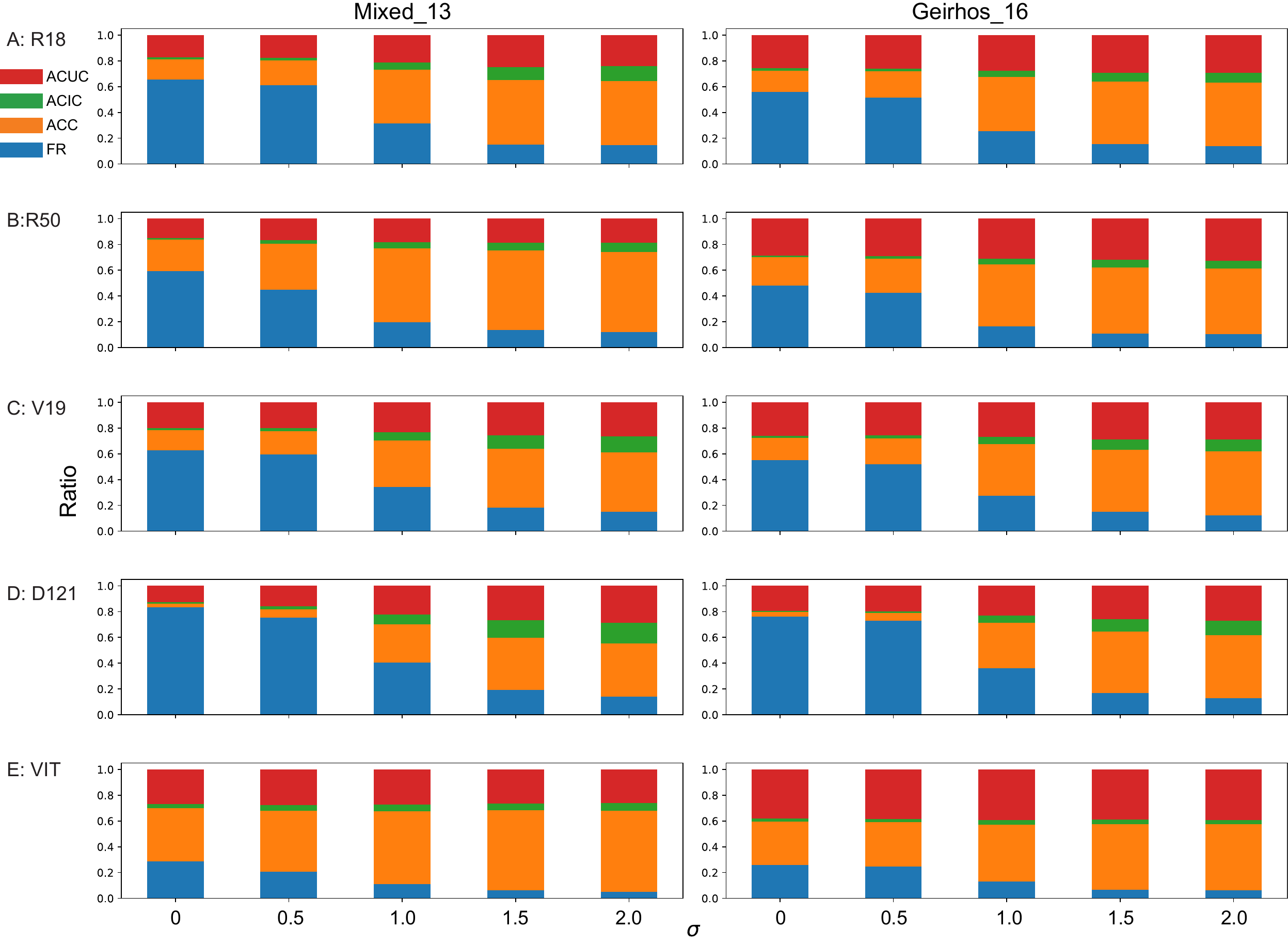}
  \caption{Evaluation of ToT on adversarial inputs crafted by AutoAttack attack. (A)-(E), evaluation of ResNet18, ResNet50, VGG19, DenseNet121 and ViT, respectively. We estimate failure ratio (FR) of detection, the accuracy of final answers (ACC), the rate of inaccurate final answers (ACIC) and the rate of inputs(ACUC), on which ToT refuses to make prediction. They are shown in blue, orange, green and red, respectively. We show evaluation on Mixed\_13 in the left column and those on Geirhos\_16 in the right column. }\label{autoattack}
\end{figure}

\begin{figure}[h]
  \centering
  \includegraphics[width=5.0in]{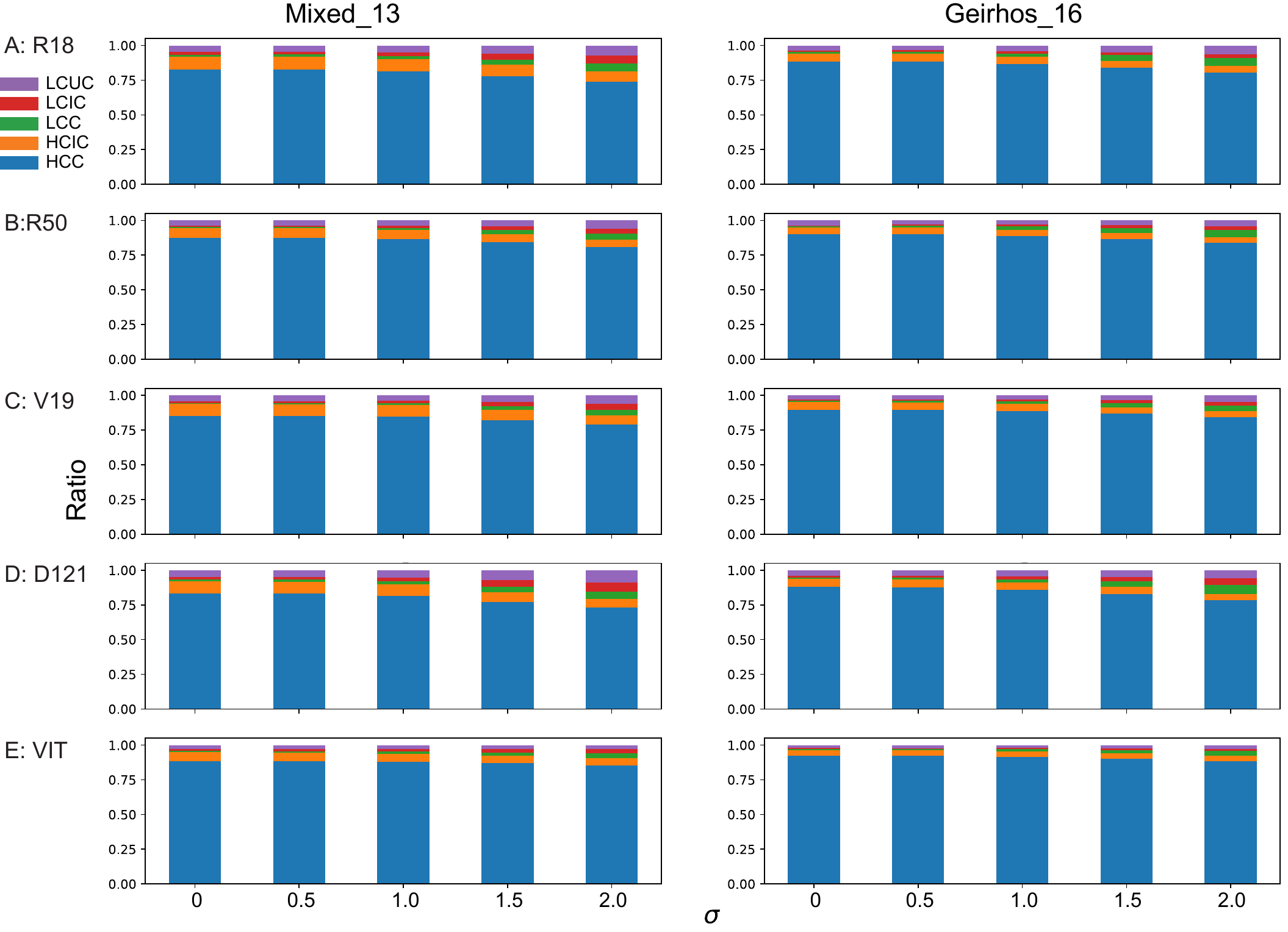}
  \caption{Evaluation of ToT on normal inputs. (A)-(E), evaluation of ResNet18, ResNet50, VGG19, DenseNet121 and ViT, respectively. We show evaluation on Mixed\_13 in the left column and those on Geirhos\_16 in the right column. See the text for details.}\label{norm}
\end{figure}



\end{document}